\documentclass[runningheads]{llncs}
\usepackage{graphicx}
\usepackage{booktabs}
\usepackage{tikz}
\usepackage{comment}
\usepackage{amsmath,amssymb} 
\usepackage{color}
\usepackage{multirow}
\usepackage{float}
\usepackage{url}
\usepackage{caption}
\usepackage{subfigure}

\begin{document}
\pagestyle{headings}
\mainmatter
\def\ECCVSubNumber{3151}  

\title{Semantic-guided Multi-Mask Image Harmonization} 

\titlerunning{Semantic-guided Multi-Mask Image Harmonization}
%
\author{Xuqian Ren\inst{1}\orcidID{0000-0002-3811-0235} \and
Yifan Liu\inst{2}\orcidID{0000-0002-2746-8186}}
%

\institute{Beijing Institute of Technology \and 
University of Adelaide}
\maketitle

\begin{abstract}
        Previous harmonization methods focus on adjusting one inharmonious region in an image based on an input mask. They may face problems when dealing with different perturbations on different semantic regions without available input masks. To deal with the problem that one image has been pasted with several foregrounds coming from different images and needs to harmonize them towards different domain directions without any mask as input, we propose a new semantic-guided multi-mask image harmonization task. Different from the previous single-mask image harmonization task, each inharmonious image is perturbed with different methods according to the semantic segmentation masks. Two challenging benchmarks, HScene and HLIP, are constructed based on $150$ and $19$ semantic classes, respectively.
        Furthermore, previous baselines focus on regressing the exact value for each pixel of the harmonized images. The generated results are in the `black box' and cannot be edited. In this work, we propose a novel way to edit the inharmonious images by predicting a series of operator masks. The masks indicate the level and the position to apply a certain image editing operation, which could be the brightness, the saturation, and the color in a specific dimension. The operator masks provide more flexibility for users to edit the image further. Extensive experiments verify that the operator mask-based network can further improve those state-of-the-art methods which directly regress RGB images when the perturbations are structural. Experiments have been conducted on our constructed benchmarks to verify that our proposed operator mask-based framework can locate and modify the inharmonious regions in more complex scenes. Our code and models are available at \url{https://github.com/XuqianRen/Semantic-guided-Multi-mask-Image-Harmonization.git}.
\end{abstract}

\section{Introduction}

Image editing techniques play a more crucial role in our daily life. They have been widely used in advertisement propaganda, social media sharing, and digital entertainment. Furthermore, with the rapid expansion of electronic devices and image processing applications, such as PhotoShop and MeituPic, image composition has become a more accessible technique. However, without professional photo-editing experience, a fake composited image will have lower evaluation credibility due to its inharmonious color, texture, or illumination. Thus, image harmonization is an imperative process to improve composite image quality. 


Previous image harmonization aims to improve the quality of image composition by matching the appearance between a pre-defined foreground region and background to make the whole composited image more realistic. The adjusted appearance can be color, brightness, contrast, etc. Deep-learning based methods~\cite{cong2020dovenet,cong2021bargainnet} usually handle this task as an image-to-image translation task~\cite{isola2017image,liu2018auto}. These methods have two limitations. First, these networks focus on the problem that one composite image only has one inharmonious region to change and do not consider the situation that the composite image may have multi-inharmonious regions with different kinds of perturbations but without masks to indicate. In a real application, the user may cut out the portraits from different photos and paste them on the same background. Second, the generated result is not explainable and editable, which is not flexible for users to further edit if they do not like the results generated by the automation output.

To solve these drawbacks, we propose a new setting named Semantic-guided Multi-Mask Image Harmonization(Sg-MMH), which aims to adjust the inharmonious regions according to their semantics. We construct two new benchmarks, named HScene and HLIP, which originated from public semantic segmentation datasets ADE20K~\cite{zhou2019semantic} and LIP~\cite{gong2017look}, separately. For each image, we randomly select several masks with different semantics to apply the perturbations. For the HScene, we focus on harmonizing various things and stuff in natural scenes, such as houses, lawns, and animals. And for the HLIP, we pay attention to adjusting the body parts and clothes of a person.

%
To realize the perturbation, we use Instagram filters, LAB perturb methods, blur perturbations to enrich disturbance scenarios, and apply different perturb functions to different regions randomly. That means the framework needs to harmonize the foreground regions towards different domains, which is a more challenging task.

Furthermore, we propose a new framework that can generate Operator Masks. Different from previous works that directly regress RGB images, we choose to generate operator masks for some pre-defined operators. These operator masks can simulate the manipulation functions in the image processing software. In this work, we define six feasible operator masks, and we multiply and add every two masks to each L, A, and B channel of inharmonious input images. Hence, each mask can be seen as an operation that can change the illumination, contrast, or color. The mask location indicates where to apply the operation, and the value of each pixel can be seen as the level each function is acted. The final generated mask is also editable and explainable, and it can locate the inharmonious region automatically. Our framework can be applied to any off-the-shelf backbones to adapt them to our task and make their output editable.

We summarize our contributions in three-fold:
	\begin{itemize}
		\item[$\bullet$] We propose a new setting that aims to simulate the situation that the foregrounds come from different images and need to be harmonized towards different domains. We propose two benchmarks, HScene and HLIP, to represent the scenery and person harmonization task separately. To enrich the training set, we perturb several foreground images with multiple semantic-guided masks on the fly.
		
		\item[$\bullet$] We propose a new framework to generate operator masks to simulate image editing operations in image processing software to harmonize images instead of generating each pixel separately in the training process. The masks indicate the level and the position to apply a certain operation, which could be the brightness, the saturation, and the color in a specific dimension.

		\item[$\bullet$] Quantitative and qualitative comparisons demonstrate that the proposed operator mask-based network can have a reasonable result on our multi-mask image harmonization benchmarks. It can also provide more explainable and editable outputs for further change.

	\end{itemize}

\section{Related Works}
\label{sec:relate}

\textbf{Image harmonization}: Traditional image harmonization methods concentrated on learning and adjusting statistical appearance between hand-crafted heuristics appearance features, such as~\cite{pitie2005n,reinhard2001color,cohen2006color,jia2006drag,perez2003poisson,tao2010error,sunkavalli2010multi}. Some of them have been adapted to image editing software. However, these methods are not very reliable when dealing with more complex situations. Recently several CNN networks started to use learning methods to make the harmonized image more reasonable. Zhu \textit{et al.}~\cite{zhu2015learning} trained a discriminative model to rank the composite images according to their realism level and optimized the model by maximizing the output visual realism score. DIH~\cite{tsai2017deep} proposed the first end-to-end CNN network to produce harmonized images with 
auxiliary semantics prediction. 
	
Since the remarkable effect of GAN in image-to-image translation~\cite{isola2017image,mirza2014conditional,zhu2017unpaired}, some GAN frameworks have also been proposed to facilitate image harmonization. 
$S^{2}AM$~\cite{cun2020improving} considered the harmonization task as just changing the foreground appearance by keeping semantic information. So it used channel-wise and spatial-wise attention mechanisms to focus on the appearance changes in the foreground mask region. DoveNet~\cite{cong2020dovenet} further conceived the image harmonization task as a domain transfer problem and hypothesized that the foreground and background of an inharmonious image are captured in different conditions. So, they chose to select the foreground objects in the datasets and perturbed their color or illumination to simulate an inharmonious composite, and trained the network to regress close to the original input image in a GAN framework with a domain verification discriminator. This is a good start to deeply researching harmonization tasks with a large dataset. Other methods also re-defined the image harmonization task to make this task more practical. BarginNet~\cite{cong2021bargainnet} formulated image harmonization as background-guided domain translation. They used a domain code extractor to extract foreground and background domain code and use background domain code to guide the foreground domain during translation. RainNet~\cite{ling2021region} treated image harmonization as a style transfer problem and extracted background style information, and applied it to the foreground. D-HT~\cite{guo2021image} used the strong power of Transformer to model long-range context information to perceive information for better harmonization. 
These methods all consider that only one foreground inharmonious region needs to be edited and directly generate and regress each pixel towards the target, which lacks interpretability. Instead, in our work, we resolve the image harmonization task using preset image editing operations to edit the inharmonious images. Specifically, we predict a series of operator masks to be applied to the inharmonious input. To tackle our proposed setting, we construct two benchmarks with semantic-guided perturbed regions that need to be adjusted in one composite image.

\textbf{Photo retouch}: Photo Retouch aims to learn an automatic network to correct a raw image to a destinate expert-retouched one to enhance image quality. UEGAN~\cite{ni2020towards} learned in an unsupervised fashion to realize image-to-image mapping.
Deep Preset~\cite{ho2021deep} realized color transfer by predicting various presets in a labeled reference photo and applied them to another image to blend and retouch a photo. Their filters are supervised and preset, however, in our framework, our filters are unsupervised and we only supervise the final harmonization output, so the filter style is not static and can be defined by users. Inspired by Deep Preset, we predict operator masks for pre-defined operations implicitly. The network can predict the position and level that each operation is applied to the picture and generate operator masks to simulate the manipulations.


\section{Methods}
	\label{sec:methods}
	
	\subsection{Overview}
   Most image harmonization tasks try to solve the problem that a composite image comes from two pictures captured in different situations, which is a simple scene in image composition. In a more general case in our daily life, users may paste several foregrounds to one background image to make the picture meet their expectations. For example, users may cut the picture of a skirt from one picture, cut the picture of a hat from another picture, and then stick them to their portrait image. In this situation, the composite image needs to be changed in different directions to adjust the whole foreground harmonious with the background portrait. So in order to solve the problem, we propose a new setting to simulate this situation, named Semantic-guided Multi-Mask Image Harmonization(Sg-MMH) task, and construct two benchmarks, HScene and HLIP, for this task.
    

	\begin{figure*}[t]
		\centering	
		\includegraphics[width=\linewidth]{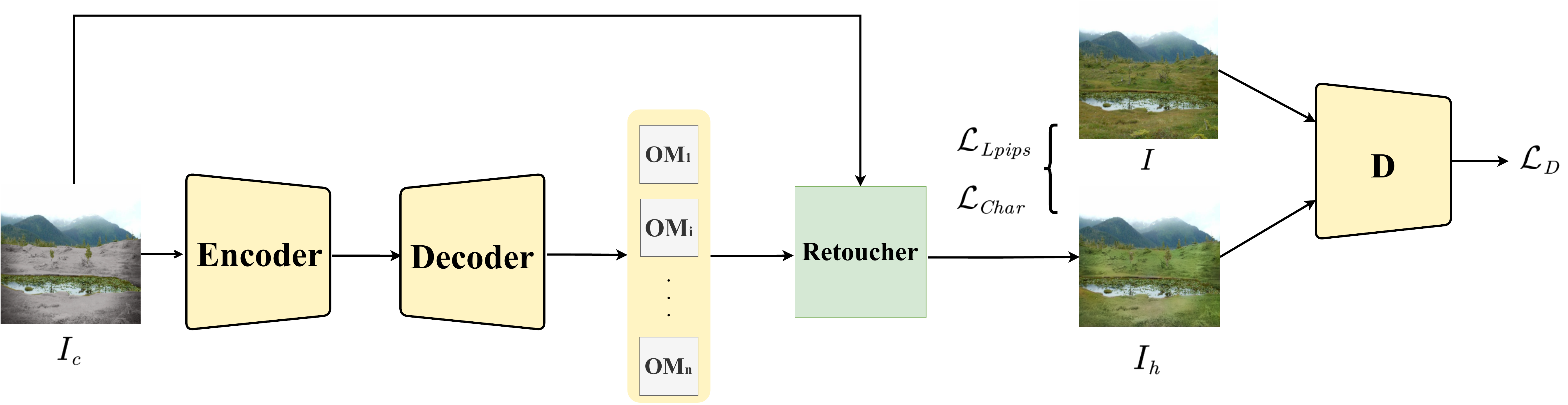}
		\caption{This image shows the training pipeline of our framework. We first generate a composited image $I_c$ guided by multiple semantic masks $M$ from the original real image $I$, serving as the input image. The encode-decode structure generates operator masks. Then the operator masks $OM$ will be used to retouch the composited image $I_c$ and formulate $I_h$. The discriminator will compare $I_h$ and $I$ to make the output $I_h$ harmonious.}
		\label{fig:structure}
	\end{figure*}

   	As shown in Figure~\ref{fig:structure}, our framework includes one generator, which contains an encoder-decoder structure and a retoucher, and a patch discriminator. First, the inharmonious image $I_c$ is the input of our operator mask generator. $I_c$ can be generated by perturbing selected regions according to the semantics of a natural or person image $I$ as described in Section~\ref{sec:data}. The encoder in the generator can locate the inharmonious region automatically. The decoder aims to predict the operator masks according to the gap between the inharmonious region and natural image distributions. These two structures can be implemented by any off-the-shelf backbone used in image harmonization tasks in previous research, such as~\cite{cong2020dovenet,cong2021bargainnet,ling2021region,guo2021image}. The output operator masks from the decoder will pass a retouch module, in which the predefined operators are applied to $I_c$ according to the predicted locations and levels to harmonize the composited image $I_c$ and generate a harmonious image $I_h$. We supervised the network by minimizing the difference between $I_h$ and the ground truth harmonious image $I$. A binary regression loss ($\mathcal{L}_{Char}$) and a perceptual loss ($\mathcal{L}_{Lpips}$) will be applied to train the model and will be further introduced in Section~\ref{sec: loss}. The discriminator we use is a patch discriminator. In this way, we force the discriminator to focus on some tiny objects and details.  
	
	In the following sections, we will first introduce how we construct our benchmarks and gradually illustrate the detailed structure of our framework and loss functions.

    \subsection{ Multi-Mask Image Harmonization}
	\label{sec:data}
	
	We define a new task called Semantic-guided Multi-mask Image Harmonization (Sg-MMH).
	The input is a composited image in which foregrounds are pasted from different pictures, and the background is another real image.
	The output is a harmonious image in which the foregrounds have been adjusted to the background.

    We build two benchmarks, \textbf{HScene} and \textbf{HLIP}. And their foregrounds have been perturbed according to their semantics. The details are as follows:

    \textbf{HScene:} The harmonized images and the semantic masks of HScene are from  ADE20K Dataset. ADE20K Dataset~\cite{zhou2017scene,zhou2019semantic} consists of natural images with high accuracy pixel level human annotations. There are 150 semantic classes, such as animals, bulidings, sky, trees, etc. There are 20k/2k/30k images in the training/ validation/ testing set.  Since the ground truth semantic label of the testing set is not released, we use the total images in the train set to construct our train set and use the validation set as our test set, and neglect some images with no foreground to construct HScene.

	\textbf{HLIP:} HLIP constructed on the basis of The Look into Person(LIP) dataset, which is a large-scale dataset that focuses on the semantic understanding of a person. It contains 50462 images with elaborated pixel-wise annotations of 19 semantic human part labels, such as dress, coat, hair and face. The original training set contains 30462 images, and the validation images contain 10k pictures. Since the images contain poses, views, occlusions, appearances, and resolution variations, they can simulate real-world portrait scenarios. To construct HLIP, we also perturb the original train set as our training set, perturb the validation set as our test set, and neglect some images without proper foreground.
	
	The statistics of training and testing images in the original datasets and our two constructed benchmarks are shown in Table \ref{tab:ratio}.

    \begin{table}
		\centering
		\renewcommand\arraystretch{1.1}
		\setlength{\tabcolsep}{1.8mm}
	    \caption{The statistics of the original datasets and our two constructed benchmarks.}
	    		
		\begin{tabular}{c|cc|cc}
			\hline
			Benchmark name & ADE20K & LIP & HScene & HLIP \\ \hline
			Training set & 20210& 30462 & 20196 &  30385 \\ \hline
			Testing  set &  2000& 10000 & 2000 &  9972 \\ \hline
		\end{tabular}

		\label{tab:ratio}
	\end{table}

    Previous works often use fixed fake foregrounds during training. So in this work, we randomly select $0.2 \times len(classes)+1$ perturb regions as the foreground and disturb them with several structural perturbations. The $classes$ number depends on the total semantic regions in each image.
    First, we mainly use Instagram filters\cite{pilgram}, which is a set of camera filters usually used in image-editing software. We implement them by a python API called pilgram. Here we use the total of 23 Instagram filters and randomly select one to perturb one region. Meanwhile, we superimpose a filter from pilgram.css to each region with a probability of 0.5. So in each benchmark, we apply different disturbations to multiple regions randomly. These Instagram filters can represent most of the common filters in image processing applications, and most of them can change the color, illumination, and saturation of the foreground structurally. 
    Second, to enrich the perturbation methods, we also add some other perturb methods to add more scenarios.  
    One is to simulate the situation that a high-resolution foreground is pasted on a low-resolution background.
    The blurring process is conducted by using several blur or noise methods, which are usually used in image super-resolution tasks to blur the real image $I$ and the background of $I_c$, and try to regress the network to blur the foreground and make the resolution of the whole image consistent. We use Gaussian noise, Laplace noise, Poisson noise, motion blur, and jpeg compression function and randomly select one blur method if blur perturbation is chosen. 
    The other additional perturb method is to randomly multiply a number $l, a,b$ to each channel of the LAB version of $I_c$ as the following equation to add more structural perturbations.
    \begin{equation}
	\centering
	I_{c_{i}}=I_{c_{i}} \times i, i=(l,a,b)
    \end{equation}

\begin{figure*}[h]
		\centering	
		\includegraphics[width=\linewidth]{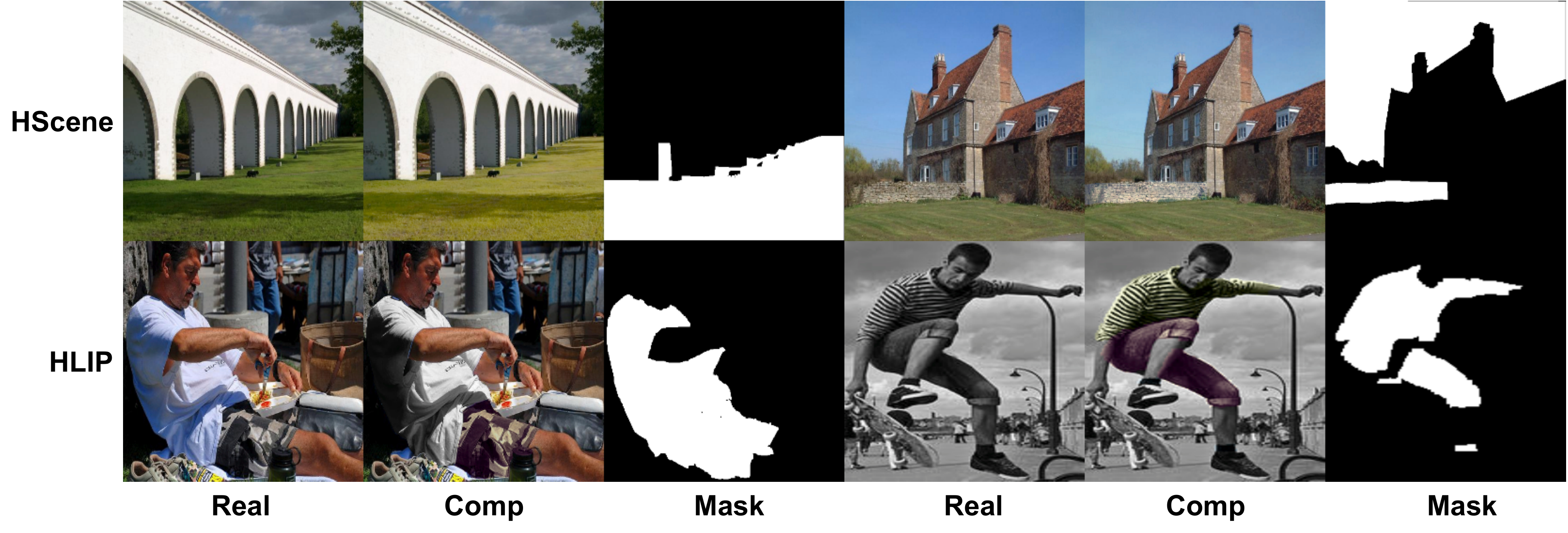}
		\caption{Here we show some perturb visualization of our benchmarks. The first line shows samples from HScene, and the second line shows examples from HLIP. We can see that the foreground color, illumination, and contrast can be changed.}
		\label{fig:perturb}
	\end{figure*}

    Some samples of our benchmarks have been shown in Figure~\ref{fig:perturb}. In this way, we want to simulate some structured perturb situation in which one region can be learned linearly in the same direction and at the same level. The reason we want to focus on this situation is that many changes in nature can be achieved by overall linear migration on a single meaningful channel, such as illumination change. So this situation is more suitable to operator masks format for they can change the appearance of one region in one direction centralized. 
    Therefore, we have different foregrounds and perturb functions every iteration, and the network needs to regress the multi-mask region in a different direction, which adds variability and complexity to each training pair.

    \subsection{Framework Structure}
    
   We first use the U-Net structure from DoveNet~\cite{cong2020dovenet} in there to illustrate our framework. It should be noted that the backbone can be replaced with any other more complex structures to boost the performance further. With our framework, the results generated in the form of operator masks are more editable and explainable.
To make the fair comparison with DoveNet~\cite{cong2020dovenet}, we replace the input of the generator as $I_c$ by removing the foreground masks, and the output of the decoder is operator masks, which its location can indicate where needs to change, and the value can represent what level needs to be changed. These operator masks could be pre-defined as certain functions and will be used to edit the inharmonious image in a Retoucher module.
    
	\textbf{Retoucher:} The operator masks will be passed into a retouch module to harmonize $I_c$. The retouch module has some preset operations which users can define. Here we employ several simple but effective preset methods. Here we show an example. We first split $I_c$ to L, A, and B channels. Then, we generate six operator masks: $OM_{i_{mul}}, OM_{i_{add}},i=(l,a,b)$, and each will be multiplied or added to each L, A, B channel of the image $I_c$, to generate $I_h$, as described in the following equation,
	\begin{equation}
		\centering
		\label{add_mul}
		I_{h_{i}}=I_{c_{i}} \odot OM_{i_{mul}} \oplus OM_{i_{add}}, i=\left( l,a,b \right) 
	\end{equation}
	
	where "$\odot$" represents the Hadamard product operation and "$\oplus$" represents the Hadamard add operation. It should be noted that other forms of retouch can also be used, and the type of the pre-defined operator can be set according to different applications or dataset volume.

	\textbf{Patch discriminator:} In our task, not only do the global features need to be discriminated, but accurate gradient feedback for local details also needs to be produced. Therefore, we employ a patch discriminator~\cite{isola2017image,ji2020real} in our framework, which has a fixed receptive field, and each output value of the discriminator is only related to the local fixed area. Therefore, we want the discriminator to focus on some tiny objects and local details in the image. The final loss is the average of all local losses to guarantee global consistency~\cite{ji2020real}.
	
	\subsection{Loss functions}
	\label{sec: loss}
	We use four type losses in our framework: 
	\begin{itemize}
		\item[$\bullet$]  We apply a binary Charbonnier loss~\cite{lai2018fast} to directly minimize the distance between our final output $I_h$ and the ground truth $I$ as:
		\begin{equation}
			\mathcal{L} _{Char}=\frac{1}{N}\sum_{i=1}^N{\sqrt{\left( I_i-I_{h_i} \right) ^2+\epsilon ^2}},
		\end{equation}
		where N is the batch size, $\epsilon$ is a very small constant in order to stabilize the value.
		
		\item[$\bullet$]  In order to enhance the contextual details, we employ perceptual loss LPIPS\cite{johnson2016perceptual}:
		
		\begin{equation}
		   	\mathcal{L}_{Lpips}=\operatorname{LPIPS}(\hat{I}, I_h)
		\end{equation}

		\item[$\bullet$]Different from standard discriminator loss, we use a relativistic discriminator loss and try to maximize the probability that a real image is relatively more realistic than a fake one~\cite{wang2018esrgan}. The discriminator loss is defined as:
		\begin{small}
		\begin{equation}
			\begin{aligned}
				\mathcal{L}_{D} =&-\mathbb{E}_{I}\left[\log \left(D\left(I, I_{h}\right)\right)\right]\\&-\mathbb{E}_{I_{n}}\left[\log \left(1-D\left(I_{h}, I\right)\right)\right] \\
				D\left(I, I_{h}\right) =&\sigma\left(C(I)-\mathbb{E}\left[C\left(I_{h}\right)\right]\right) \rightarrow 1 \\
				D\left(I_{h}, I\right) =&\sigma\left(C\left(I_{h}\right)-\mathbb{E}[C(I)]\right) \rightarrow 0
			\end{aligned}
		\end{equation}
		\end{small}
		$\sigma$ is the sigmoid activation layer, and $C(x)$ is the output of the discriminator~\cite{wang2018esrgan}.
		$\mathbb{E}_{I_h}[\cdot]$ represents the final loss is averaged on the batch-level~\cite{wang2018esrgan}.

		\item[$\bullet$]  The adversarial loss for the generator is as following~\cite{wang2018esrgan}:
		\begin{small}
		\begin{equation}
			\mathcal{L} _{G\,\,}=-\mathbb{E} _I\left[ \log \left( 1-D\left( I,I_h \right) \right) \right] -\mathbb{E} _{I_h}\left[ \log \left( D\left( I_h,I \right) \right) \right] 
		\end{equation}
		\end{small}
		In this way, the generator benefits from the gradients from both outputs harmonized image and real image in adversarial training. 
	\end{itemize}

	Finally, our total loss function used for the generator is :
	\begin{equation}
		\mathcal{L} _{total}=\alpha \mathcal{L} _{Char}+\beta \mathcal{L} _{\mathrm{Lpips}}+\gamma \mathcal{L} _{G},
	\end{equation}
	where $\alpha$, $\beta $, $\gamma$ are set as 1, 1, 0.005 empirically.

	\section{Experiments}

	In this section, we introduce implementation details. Then we evaluate our proposed framework on our constructed datasets.

	\subsection{Implementation}
	\label{implementation}
	As for the structure, in our framework, we first apply the backbone of an attention U-Net from DoveNet~\cite{cong2020dovenet}. The pre-defined operator is set to the form of formula ~\ref{add_mul}.
	So the input channel of the backbone is 3, the output channel is 6, and the number of filters in the last conv layer is 64. The number of downsamplings in U-Net is set to 8. The normalization type is instance normalization.
	
	As for the optimization, we first trained the generator from scratch with the learning rate of 2e-4. The milestone is set as 20k, 30k, and the total iteration is 40k for the HScene benchmark. And the milestone is set as 55k, 75k, and the total iteration is 85k for the HLIP benchmark. When training the whole framework with discriminator, we loaded the pre-trained generator model and trained with another 10k for HScene and HLIP. We used the ADAM optimizer, and the batch size in each GPU is set to 64. The whole training process is on Nvidia 3090 GPUs. We implemented it based on BasicSR~\cite{wang2020basicsr} framework. We used 256 $\times$ 256 resolution for both training and testing evaluation.
	
	The metric we use is Mean square Error(MSE), Peak Signal-to-Noise Ration (PSNR), Structural SIMilarity (SSIM)~\cite{wang2004image} and Perceptual metrics (LPIPS)~\cite{zhang2018unreasonable}. PSNR and SSIM pay more attention to the fidelity of the image rather than visual quality~\cite{ji2020real}. The larger the two metrics, the less likely the image is distorted. LPIPS mainly focuses on comparing the distance of visual features between the generated image and the real one. In this work, we use pre-trained VGG to extract image features, and the smaller the LPIPS is, the generated image is closer to the real one visually.

	
	\subsection{Comparison to recent works}
	
	In this section, we compare our framework with the original output, which directly regresses the RGB value of each image in both quantitative and qualitative methods. We also conduct a user study test to demonstrate the subjective effect of these algorithms.
	
	\noindent \textbf{Quantitative Comparison:}
	We evaluate the performance of our framework on our conducted two benchmarks, HScene and HLIP. The baseline network is adjusted from Dovenet\cite{cong2020dovenet}. We re-implemented it on the proposed multi-mask dataset by removing the input mask. The quantitative results are shown in Table~\ref{tab:compare}. The first line shows the metrics of the composited images on test datasets. When directly outputting the RGB version of the harmonious image, the result cannot be interpretable and modifiable, and the results are also sub-optimal in our setting. Our framework makes an improvement in terms of all evaluation metrics on two benchmarks.
	
	To make a fair comparison with other SOTA methods, we also conduct experiments on previous popular single-mask benchmarks, such as iHarmony4~\cite{cong2020dovenet}. We train and test on iHarmony4~\cite{cong2020dovenet} datasets and show the results in the supplementary materials. The experiment results demonstrate that our methods can also have a reasonable effect on the single-mask image harmonization task. 
	

    \begin{table}[t]
    \centering
    \caption{The comparison of the composite, RGB format output, and our framework output. Our framework can have a reasonable improvement on both the two benchmarks.}
    \resizebox{\textwidth}{!}{%
    \begin{tabular}{c|cccc|cccc}
    \hline
    \multicolumn{1}{c|}{Method}            & \multicolumn{4}{c|}{HScene}   & \multicolumn{4}{c}{HLIP}      \\ \hline
    \multicolumn{1}{c|}{Metric}            & MSE$\downarrow$    & PSNR$\uparrow$  & SSIM$\uparrow$ & LPIPS$\downarrow$ & MSE$\downarrow$    & PSNR$\uparrow$  & SSIM$\uparrow$ & LPIPS$\downarrow$  \\ \hline \hline
    \multicolumn{1}{c|}{Composite}         & 271.89 & 27.83 & 0.96 & 0.027 & 111.95 & 30.99 & 0.96 & 0.021 \\ \hline
    \multicolumn{1}{c|}{Sg-MHH(RGB)}       & 141.75 & 28.73 & 0.95 & 0.026 &  75.45 & 31.34 & 0.95 & 0.025    \\ \hline
    \multicolumn{1}{c|}{Sg-MHH(OM)}        & 99.84  & 30.59 & 0.96 & 0.020 & 48.02  & 33.11 & 0.96 & 0.021  \\
    \multicolumn{1}{c|}{Sg-MHH(OMGAN)}     & \textbf{94.00}  & \textbf{30.88} & \textbf{0.96} & \textbf{0.020} & \textbf{42.62}  & \textbf{33.92} & \textbf{0.97} & \textbf{0.018}\\ \hline
    \end{tabular}%
    }
    
        \label{tab:compare}
    \end{table}

\begin{figure*}[t]
		\centering	 
		\includegraphics[width=\linewidth]{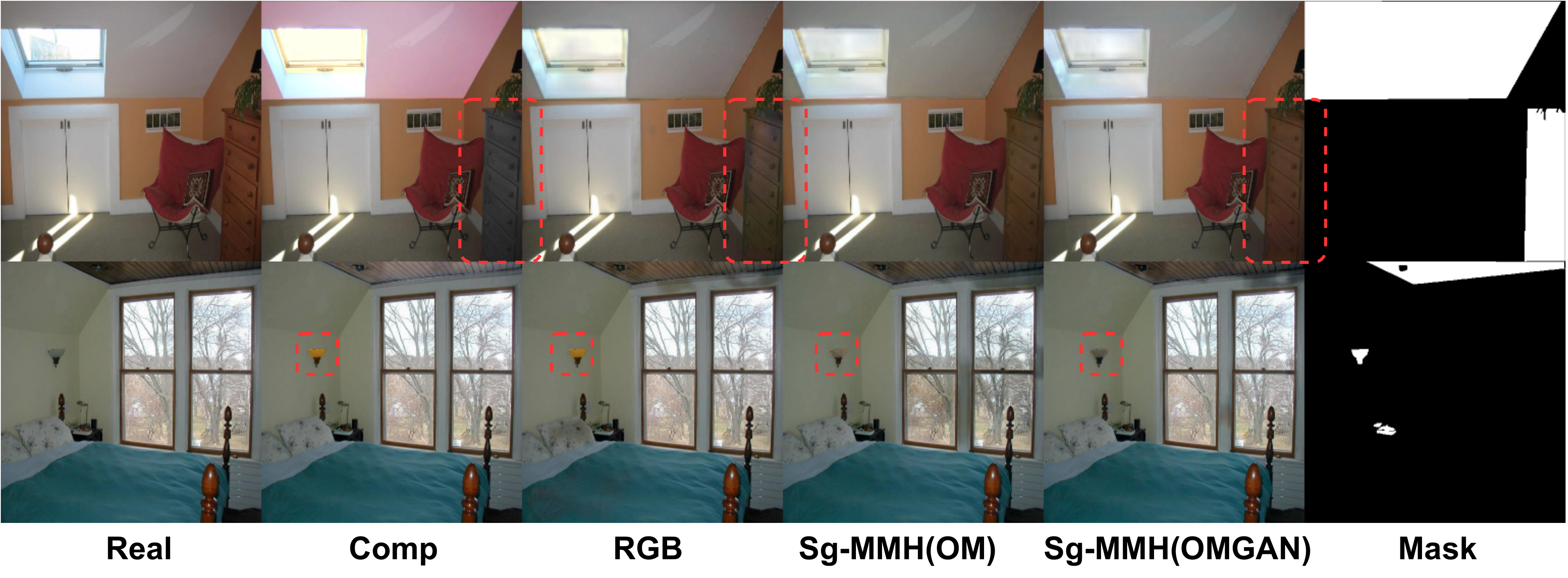} 	
		\caption{\textbf{Qualitative harmonization results.} The first column is the real image. The second column is the composited image. The third to the fifth column is the harmonization results of a baseline RGB format and our framework with or without GAN structure. OM means \textit{only use generator in our framework}. OMGAN represents \textit{generate operator masks with GAN framework}.}
		\label{fig:baseline}
\end{figure*}

\noindent \textbf{Qualitative Comparison:} We show visual comparison results on our proposed multi-mask harmonization datasets in Figure~\ref{fig:baseline} and show more visualization on supplementary. The first column shows the real images, and the second column shows the perturbed images. In the third to fifth columns, we show the results output in the previous version and the results output from our framework without or with a discriminator. As the perturb methods on different segmentation masks are different, our framework can deal with the situation with different adjustment directions properly. 
As demonstrated in the red dotted box, the methods that directly output the RGB format only adjust one inharmonious region (the results in the last line) or adjust two areas but are not in place(the results in the first line). In this way, we demonstrate that our framework has a more flexible and realistic effect and can handle more complicated scenes. 

   \noindent \textbf{User Study:}
	We conduct user study to compare the results of our framework which the backbone is from DoveNet with the original DoveNet\cite{cong2020dovenet}, BarginNet\cite{cong2021bargainnet} and RainNet\cite{ling2021region} methods and show the results in Table~\ref{tab:user}. We test all methods on real composited images. 
	
	\begin{table}[t]
		\begin{center}
		\caption{\textbf{User study results on real composited images.} We show four images to users and ask them to choose the most realistic one. The number means what percentage of subjects think the images produced by one method are more realistic.}
			\begin{tabular}{p{4cm}|c}
				\hline
				Methods & User Study Score(\%)$\uparrow$ \\ \hline
				DoveNet\cite{cong2020dovenet}      &20.6  \\ 
				BarginNet\cite{cong2021bargainnet} & 24.2 \\ 
				RainNet\cite{ling2021region}       & 18.5  \\\hline
				Sg-MMH(OMGAN)                              & 36.7   \\ \hline
			\end{tabular}
			\label{tab:user}
		\end{center}
	\end{table}
	
	\begin{figure}[h]
		\centering	 
		\includegraphics[width=\linewidth]{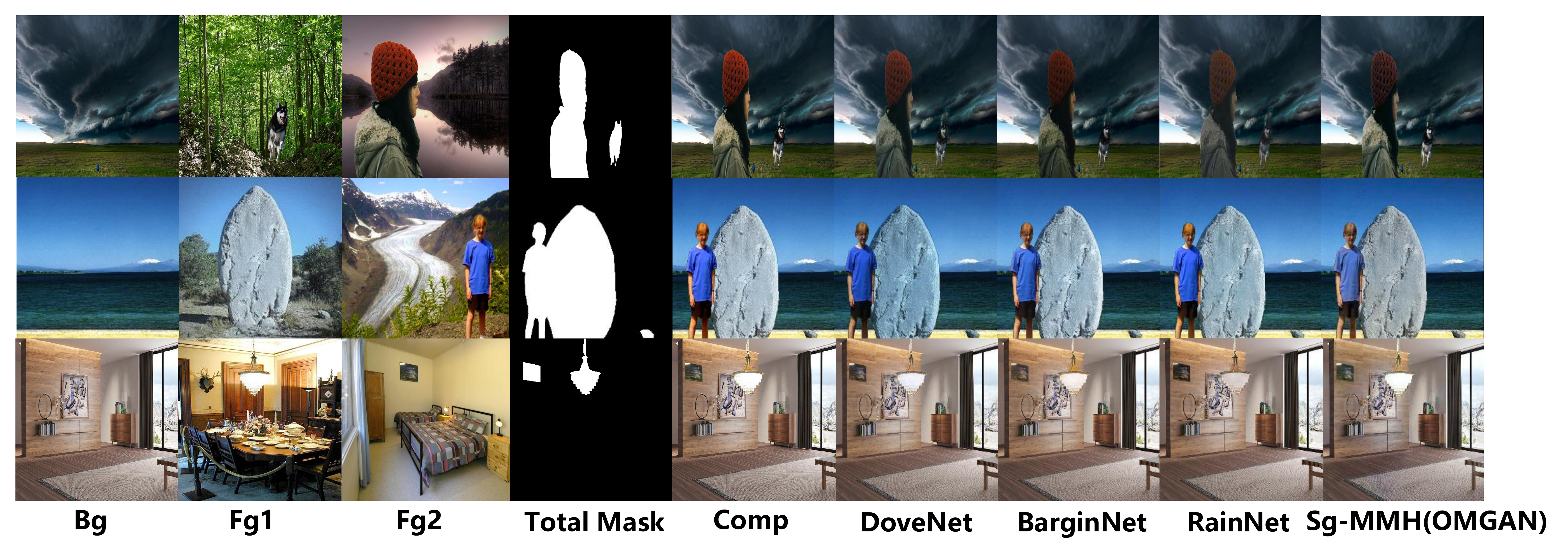}	 	
		\caption{\textbf{Qualitative results on real composited images}. We compare our method with recent state-of-the-art harmonization methods, including DoveNet, BarginNet, and RainNet. Note that they all have input masks to indicate the inharmonious regions while our method adjusts the input automatically. Our proposed framework can apply different operations for different semantic regions appropriately.}
		\label{fig:qualitative}
	\end{figure}
	We consider three realistic situations to construct real composited images. 1) Real composite images that need structured adjustment, such as only changing the illumination. 2) Real composite images that foregrounds are cut from pictures that have already been processed by Instagram filters and want to be re-processed. 3) Real composite images that the foregrounds come from two real pictures but already look realistic. Both of the composited images contain multiple regions that need to be changed.
	We construct 11 image pairs showing one picture of each setting in Figure~\ref{fig:qualitative}.
	Since there is no real image as 
    reference, we cannot use evaluation like proposed in ~\ref{implementation} to evaluate image quality. We asked 60 subjects, and each subject needed to independently select one image they think is the most realistic of the four images.  
    The images from the four methods are presented randomly. 
    From the qualitative results and user study, we can see that, when dealing with the illumination change task, our method only change the brightness level but retain other color prosperity. This is meaningful because, in most cases, we just want to change the illumination of the foreground but keep the color of the foreground itself. 
    When dealing with pictures such as downloading from websites which already processed by camera filters, our method can alleviate these filters and change them to a more natural appearance. This situation is also worth paying attention to because, in many cases, the landscape images posted to the Internet have already been processed with camera filters, but traditional color transfer methods cannot mimic the realistic filters. 
    When dealing with pictures in which the foregrounds are cut from real pictures, our model mostly keeps remaining the foreground color because although the foregrounds and background are not from the same picture, the original foreground is also reasonable in the new environment. In many cases, when the foreground and background colors are not very consistent, the composite image also looks actual and sensible and does not necessarily need to be adjusted. Only when the color difference is very large and unreasonable the harmonization task needs to adjust the composite image appropriately.

     \begin{figure}[t]
		\centering	 
		\includegraphics[width=\linewidth]{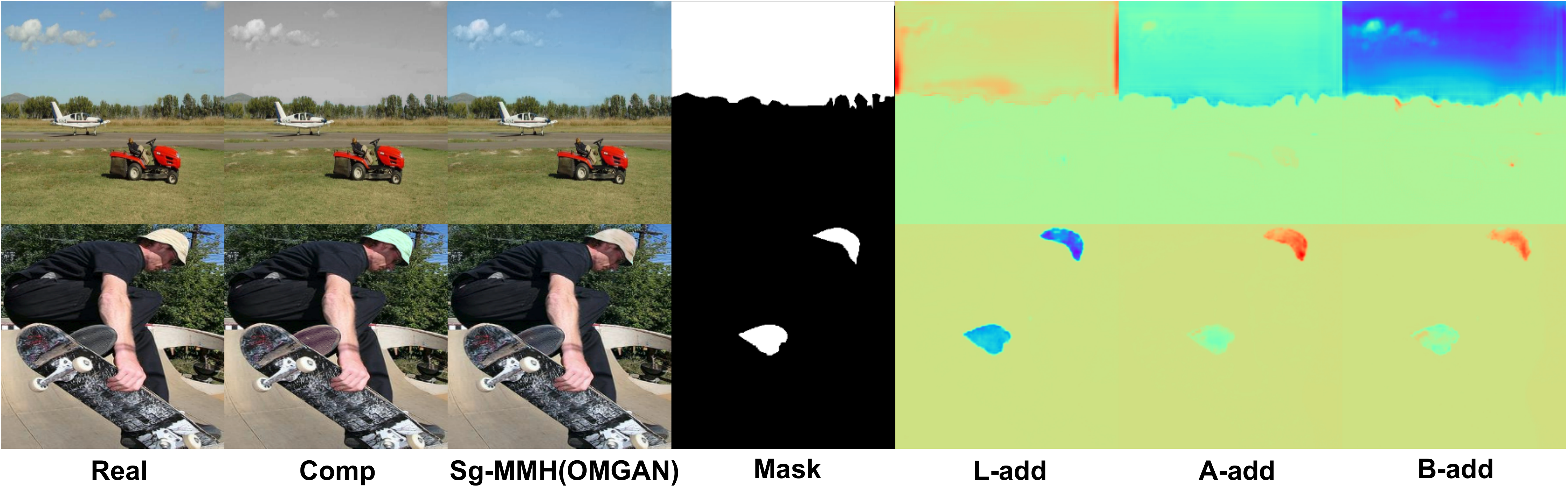}	 	
		\caption{\textbf{Visualization of the operator masks in our datasets.} The first line is the result from the HScene dataset, and the last line is the result from the HLIP datasets. The last three columns visualize the operator masks of `add' of L, A, and B channels.}
		\label{fig:visualization}
	\end{figure}

\subsection{Discussion}
 \noindent\textbf{Explainable property of operator masks:} 
  Our operator masks can sense the changes and respond appropriately to the corresponding operator masks. Here in Figure~\ref{fig:visualization} we visualized our operator masks effect on our datasets. For example, in the first line, the sky is perturbed to gray, and to recover it into blue, the operator mask $OM_{b_{add}}$ has a largely negative response since when the picture is in LAB format, the smaller the value of channel B, the bluer the color. 
  In the second line, the foregrounds are disturbed in two directions, the hat is perturbed greener, and the shoes become redder, so $OM_{a_{add}}$ and $OM_{b_{add}}$ has a positive response in the hat region and has a negative response in the region of the shoe. 
    These pictures show that our algorithm can explain the changes the network makes for the inharmonious input.
	
\begin{figure}[h]
		\centering	 
		\includegraphics[width=\linewidth]{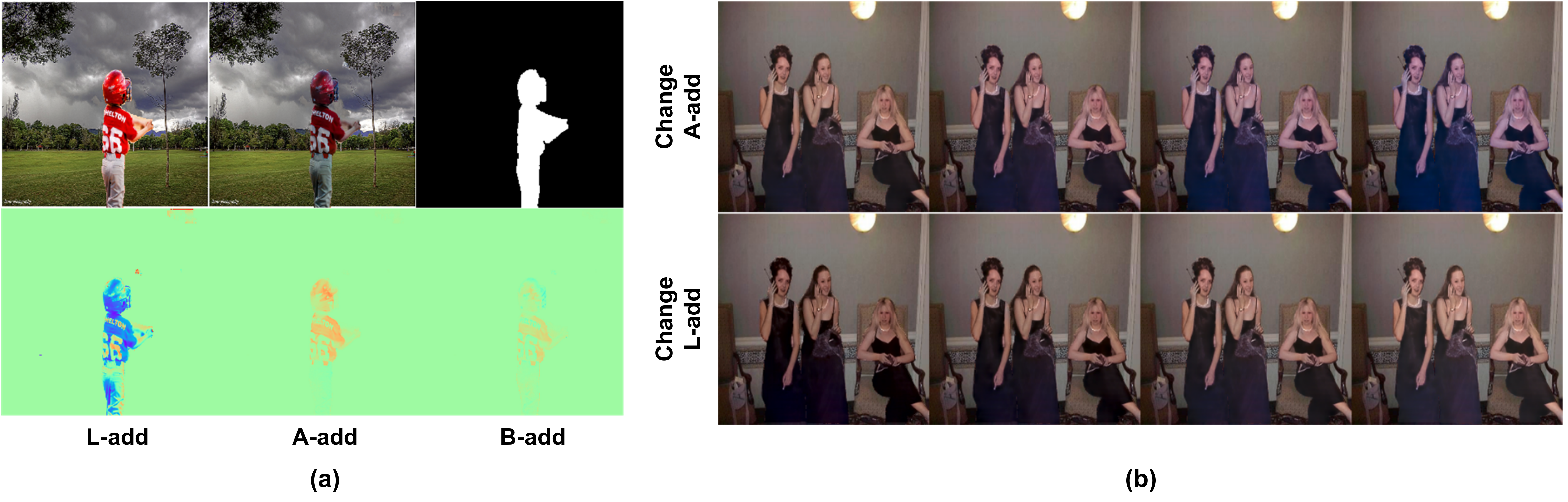}
		\caption{Here we show the explainable and editable attributes of our operator masks. Figure 5(a) shows the \textbf{Visualization of the operator masks in real data}. First row: the composite, the harmonious result, and the ground truth foreground mask. Second row: the operator masks of `add' operator masks in L, A, and B channels. The L channel has a strong negative activation, indicating the brightness of the inharmonious region has been reduced. Figure 5(b) shows the \textbf{Editing attribute of the operator masks.} First row: gradually enlarge the value of the `add' operator mask in the A channel. Second row: gradually enlarge the value level of the `add' operator mask in the L channel, which is the brightness. This figure shows that our method can edit the harmonization results based on the generated results.}
		\label{fig:attributes}
		
	\end{figure}

	To illustrate that our method can also explain when dealing with real images, we cut a foreground and paste it into a real background picture. Here in Figure~\ref{fig:attributes}(a) show that our operator mask $OM_{l_{add}}$ has a strong negative activation, indicating that the brightness of the inharmonious region is reduced.
	
	\noindent \textbf{Editable property of operator masks:} 
	Previous work often generates harmonized images directly, and the results cannot be edited if users do not like the automatic outputs. However, in our work, we locate and generate operator masks, which can be edited if users do not like the results. Here Figure~\ref{fig:attributes}(b) shows the edible effect of our algorithm. Users can add or multiply a number to deepen the color or change the illumination of the foreground. This figure shows that our method can edit the harmonization results based on the automated generation.

   \noindent\textbf{Limitation and future work:} 
   Our method attempt to first realize the harmonization framework with operator masks. Although we have explored the effectiveness of operator masks and show their editable and explainable features, there are still some limitations that need further exploration. 
   First, the training of the harmonization network may heavily depend on the perturbations. The particularity of the filters we use makes our model have many performance limitations. The performance and generalization ability of the network can be further improved if more perturbations are included in the data pre-processing. Also, without special restrictions, some filters applied in our benchmarks make some rare composited images occur. In future research, filters should be applied according to the semantics of objects, such as a green filter should not be applied to a face. 
   Second, when dealing with the foreground color, current image harmonization methods usually choose to transform the color of the foreground to be consistent with the background, but this action is often unnecessary. Sometimes the foreground can keep its own color, and the whole composite image also looks reasonable and realistic. Future work could go further step to explore in which region the foreground color can be consistent with the background and give a color range, rather than just outputting one result.
   Third, we propose six alternative operator masks to build our retoucher. However, more complicated operators require further exploration, for example, the shadow effects, the grainy effect, or de-noise.
   Forth, as our operator masks are trained in an unsupervised manner, there may be coupling between different masks. The controllability can be further improved if the operator masks can be decoupled or with specific supervision.

   

	\section{Conclusion}
	In order to tackle the problem that one composite image has multiple foregrounds pasted from other images, we set up a new setting Sg-MMH and propose two benchmarks, HScene and HLIP, to simulate the situation. To solve the benchmarks, we propose a new framework that uses operator masks to solve the multi-mask image harmonization task. The proposed methods can simulate preset operations in the PhotoShop to edit each image channel and can be explained and edited by users. Also, the operator masks can locate the inharmonious region automatically even when the indicator masks are not given. Experiments show that our framework can handle the more complicated scenes and achieve good results. There are also some limitations for the current work, such as the design of the operator masks is easy and can not decoupled well. In the future, more parametric operator masks can be applied to this framework based on expert knowledge, which will add more control and interpretability to the image harmonization task.


\clearpage
%
%
\bibliographystyle{splncs04}
\bibliography{egbib}
\end{document}


\pagestyle{headings}
\mainmatter
\def\ECCVSubNumber{3151}  

\title{Supplementary Material for Semantic-guided Multi-Mask Image Harmonization} 

\titlerunning{Semantic-guided Multi-Mask Image Harmonization}
%
\author{Xuqian Ren\inst{1}\orcidID{0000-0002-3811-0235},
Yifan Liu\inst{2}\orcidID{0000-0002-2746-8186}}
%
\institute{ Beijing Institute of Technology\\
 University of Adelaide}
\maketitle

In this supplementary material, we will first show more illustration cases about the explainable property of our proposed framework.
%
%
Second, we show more qualitative results of baseline and our proposed framework. 
Third, more visualizations in our user study will be provided.
Next, some ablation studies is conducted.
Finally, we will compare our framework with other state-of-art methods on a public benchmark with single masks.

	

\section{Explainable property of our framework}
\begin{figure*}[t]
		\centering	
		\includegraphics[width=\linewidth]{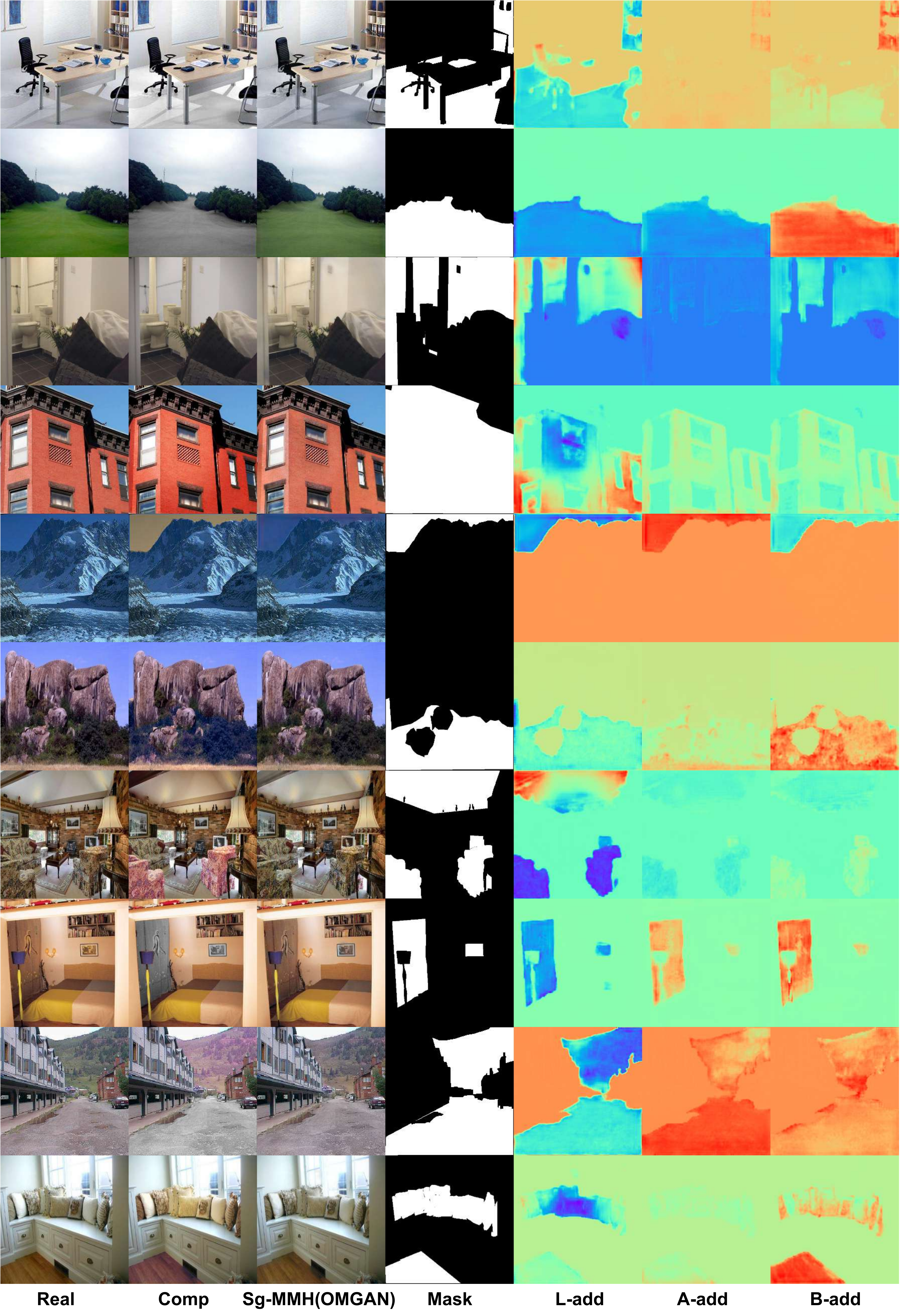}
		\caption{Here we visualize more examples from HScene. Our Operator Masks can make positive or negative responses according to different changes in the foregrounds.}
		\label{fig:HScene_vis}
	\end{figure*}

\begin{figure*}[t]
		\centering	
		\includegraphics[width=\linewidth]{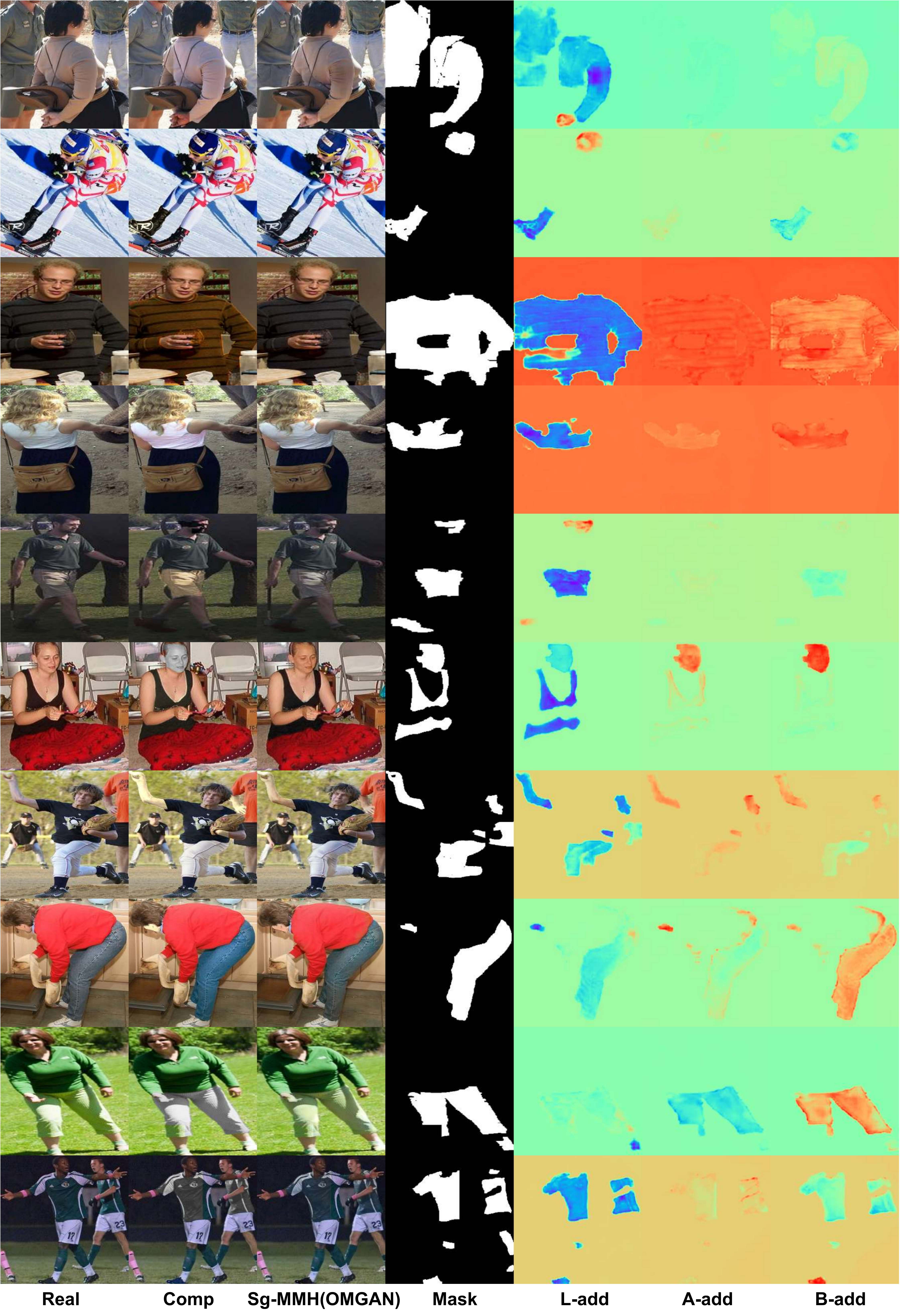}
		\caption{Here we visualize more examples from HLIP. Our Operator Masks can make positive or negative responses according to different changes in the foregrounds.}
		\label{fig:HLIP_vis}
	\end{figure*}

Here we show more results to demonstrate the explainable property of our framework by visualizing operator masks. Figure~\ref{fig:HScene_vis} are the results from HScene, and Figure~\ref{fig:HLIP_vis} shows results from HLIP. 
We can see that our framework can sense the disturbance in various directions and give a reasonable explanation for its adjustment direction. 
Thus, our method can make the harmonization process interpretability as well as ensure the performance is similar to previous baselines. 
We can also see that our framework can harmonize two completely different adjustment directions in an image (like results show in the 1, 7, 8, 9, 10\textit{th} line of Figure~\ref{fig:HScene_vis} and the 1, 2, 5, 7, 8, 9, 10\textit{th} line of Figure~\ref{fig:HLIP_vis}), which will not make the framework adjust the foreground in only one direction, so our framework is foreground-aware network.


\section{Qualitative results between different output formats}

\begin{figure*}[t]
		\centering	
		\includegraphics[width=0.9\linewidth]{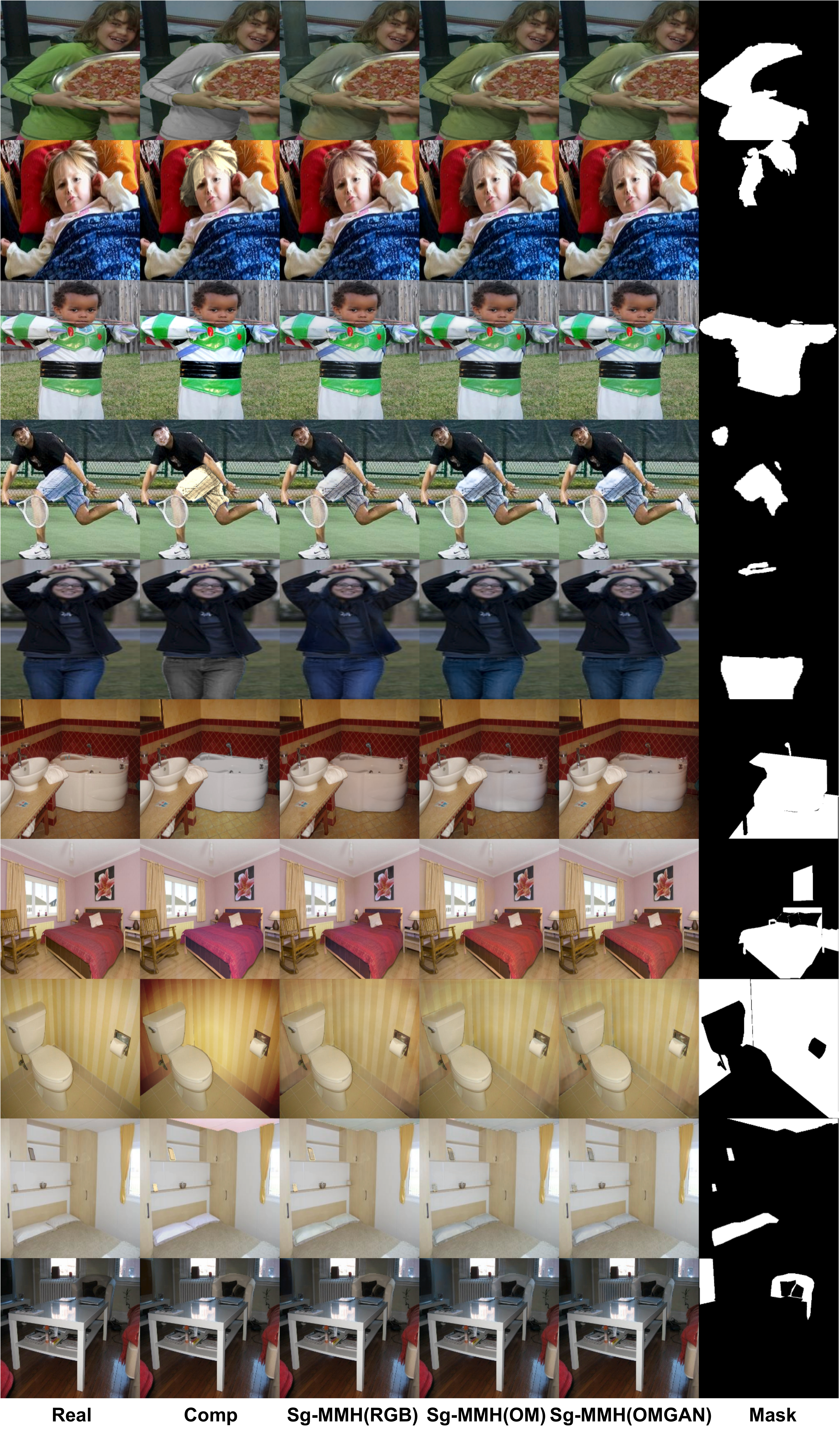}
		\caption{Here we show more comparison results between RGB output format and our Operator Mask output w/wo discriminator.}
		\label{fig:more_compare}
	\end{figure*}

In Figure~\ref{fig:more_compare}, we show more qualitative visualization results between the RGB output format and our operator masks output format. Our operator mask-based framework is more suitable for structural adjustment, even there are different harmonization directions in one image. Also, training in a GAN framework can make the model pay more attention to the details(as can be seen in the 9\textit{th} line in Figure~\ref{fig:more_compare}).


\section{More User study results}

\begin{figure*}[]
     
		\centering	
		\includegraphics[width=\linewidth]{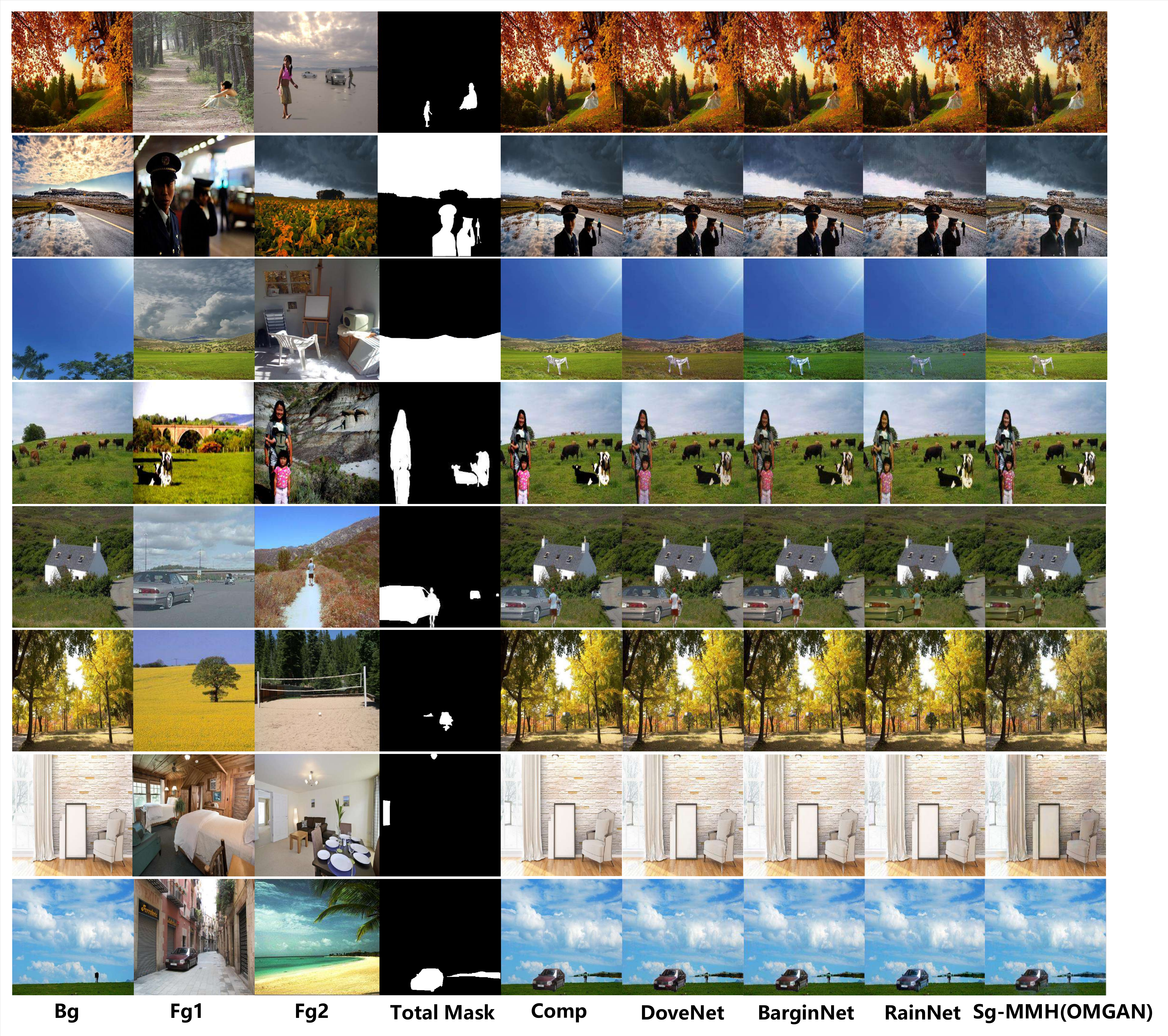}
		\caption{Here we show more visualizations used in the User Study to compare the results of our method with DoveNet\cite{cong2020dovenet}, BarginNet\cite{cong2021bargainnet}, and RainNet\cite{ling2021region}.}
		\label{fig:user}
	\end{figure*}

Here we show more pictures used in the User Study in Figure~\ref{fig:user}. The first three lines show results that only illumination needs to be changed. The 4, 5\textit{th} shows the results when foregrounds have already been processed by Instagram filters. The last few lines provide the results when foregrounds from different pictures, but the composited images already look real. So our model tends to keep them almost unchanged.


\section{Ablation study on HScene}
In this section, we show the effect of each component in our framework in Table~\ref{tab:ablation}. It can be seen that both the perceptual loss and operator masks add improvement to the final harmonious image.
In order to better study the localization of our framework, we limit the output and changes to $OM_{add}$ and binarize the operator masks. The IOU between our OMs and ground truth masks is 52.9\%.

\begin{table}[]
\centering
\caption{Effects of different components in our framework.}
\begin{tabular}{cccc|cccc}
\hline
$OM_{add}$ & $OM_{mul}$ & $\mathcal{L}_{Lpips}$ & $D$ & MSE$\downarrow$ & PSNR$\uparrow$ & SSIM$\uparrow$ & LPIPS$\downarrow$ \\ \hline
 &  & $\checkmark$ &  & 141.75 & 28.73 & 0.95 & 0.026 \\
$\checkmark$ &  & $\checkmark$ &  & 92.45 & 30.79 & 0.96 & 0.052 \\
$\checkmark$ & $\checkmark$ & $\checkmark$ &  & 99.84 & 30.59 & 0.96 & 0.020 \\
$\checkmark$ &$\checkmark$ &  &  & 105.27 & 30.29 & 0.96 & 0.021 \\
$\checkmark$ & $\checkmark$ & $\checkmark$ & $\checkmark$ & 94.00 & 30.88 & 0.96 & 0.020 \\ \hline
\end{tabular}

\label{tab:ablation}
\end{table}


\section{Two-Stage v.s One-Stage}
Our framework can harmonize regions without input masks, integrating localization and harmonization functions. We compare our one-stage pipeline with a two-stage pipeline. Previous methods need masks as input. When masks are not provided, the localization method needs to be first used to get the inharmonious region. We first use DIRL\cite{9428309} to locate the inharmonious regions and then use DoveNet\cite{cong2020dovenet} to harmonize them progressively. The results can be seen in Table~\ref{tab:two-stage}. Our single-stage model greatly saves the processing time and have higher precision.
\begin{table}[]
\centering
\vspace{-0.5cm}
\caption{Comparison between the Two-Stage pipeline and the One-Stage pipeline. The time is the whole seconds to process the test images. $^\ddag$ means we fine-tune the whole model. We fine-tune DIRL 60 epochs and DoveNet 200 epochs.}
\resizebox{\columnwidth}{!}{%
\begin{tabular}{c|ccccc|ccccc}
\hline
\multirow{2}{*}{Methods} & \multicolumn{5}{c|}{HScene} & \multicolumn{5}{c}{HLIP} \\ \cline{2-11} 
 & MSE$\downarrow$ & PSNR$\uparrow$ & SSIM$\uparrow$ & LPIPS$\downarrow$ & Time(s)$\downarrow$ & LPIPS$\downarrow$ & PSNR$\uparrow$ & SSIM$\uparrow$ & LPIPS$\downarrow$ & Time(s)$\downarrow$ \\ \hline
DIRL$^\ddag$\cite{9428309}+DoveNet$^\ddag$\cite{cong2020dovenet} & 158.94  & 29.06 & 0.95 & 0.066 & 287.6  & 80.34 & 31.45 & 0.96 & 0.047 & 1430.0 \\
Sg-MMH(OMGAN) & 94.00 & 30.88 & 0.96 & 0.020 & 106.6 & 42.62 & 33.92 & 0.97 & 0.018 & 599.0 \\ \hline
\end{tabular}%
}
\vspace{-0.8cm}
\label{tab:two-stage}
\end{table}



\section{Results on iHarmony4}
    
    Our framework mainly focuses on structural perturbations and the multi-mask harmonization task. However, to further verify the effectiveness of our framework, we tested it on the existing dataset. We choose iHarmony4\cite{cong2020dovenet} to further explore the ability of our framework when it is applied to a normal single-mask harmonization task. 
    We adapted an Harmonization Transformer~\cite{guo2021image} (HT) to implement our framework and replaced the output as operators masks.The results are shown in in Table~\ref{tab:iharmony4}.
    
\begin{figure*}[t]
		\centering	
		\includegraphics[width=\linewidth]{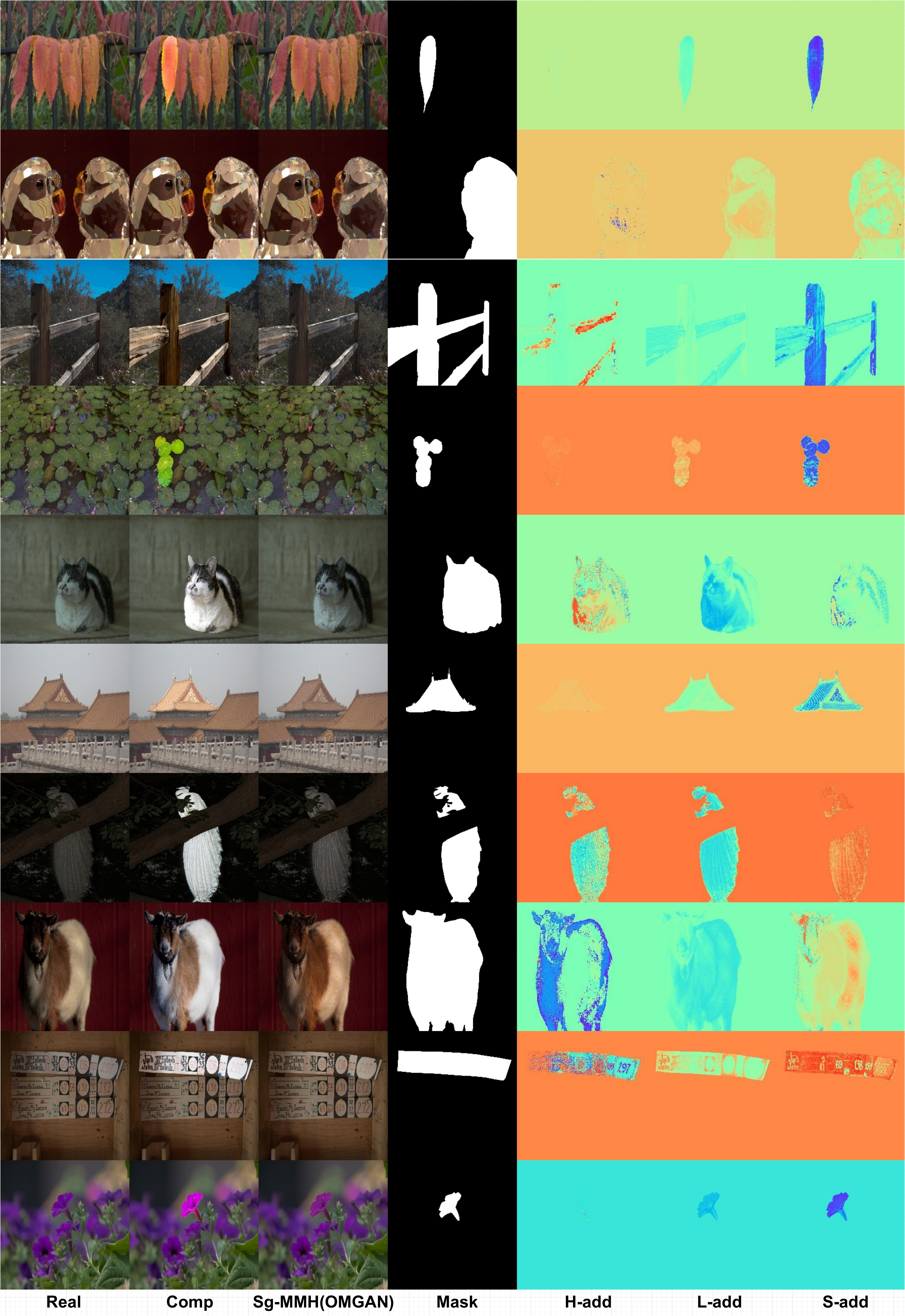}
		\caption{Here we visualize some examples from iHarmony4. Our Operator Masks can make positive or negative responses according to different changes in the foregrounds.}
		\label{fig:more_iharm}
	\end{figure*}

\begin{figure*}[t]
		\centering	
		\includegraphics[width=\linewidth]{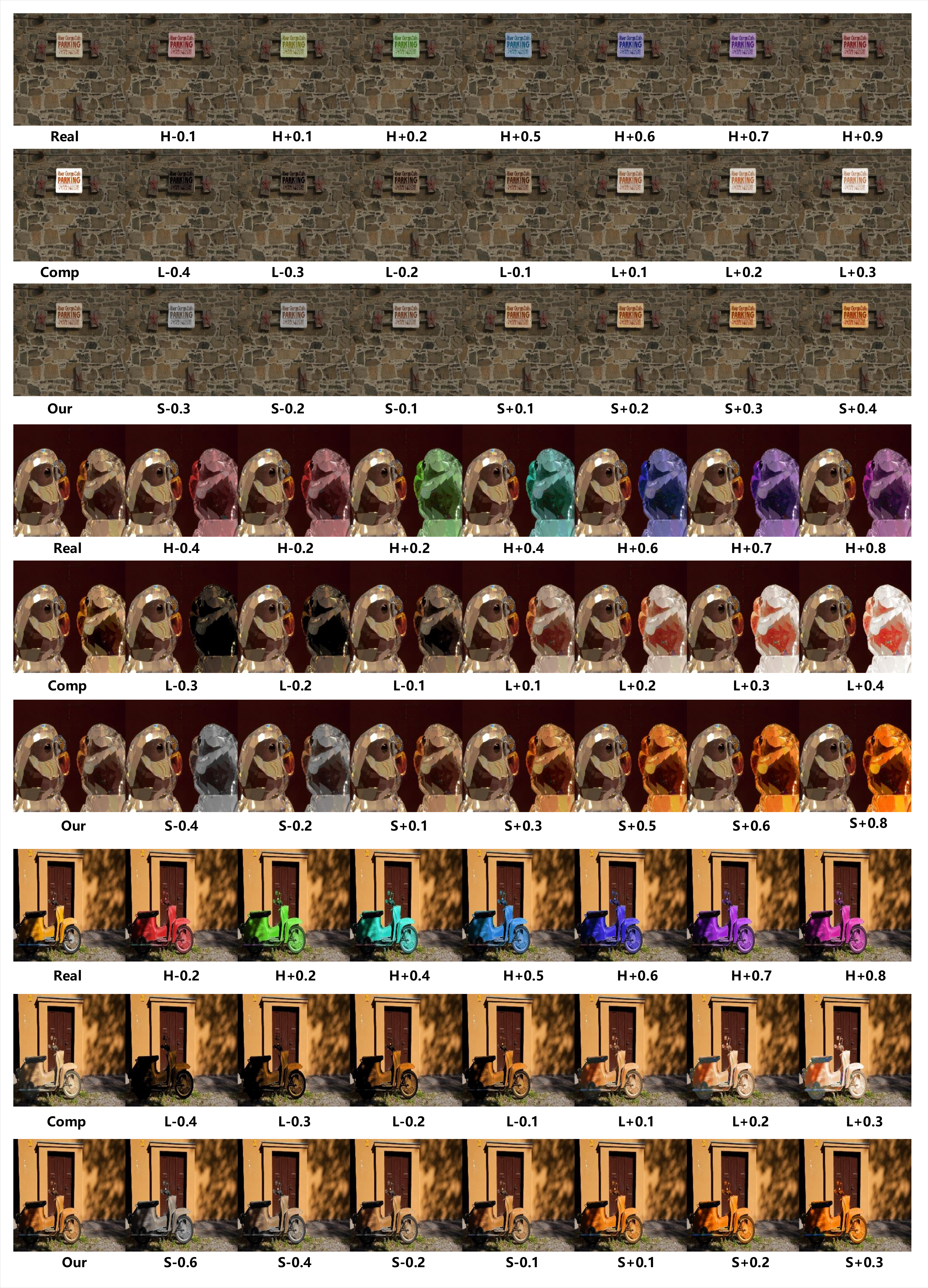}
		\caption{Here we visualize the editable property with samples from iHarmony4. We change the $OM_{add}$ in H, L, and S channels separately to gain various outputs.}
		\label{fig:more_change}
	\end{figure*}

    In order to explore the effects of more kinds of operation masks, we also implement experiments on HLS Color space, for it has more decoupled channels. The operations on the H channel can manipulate the color independently, and the operator masks on L and S channels can change the illumination and saturation individually. So it is a better space to manipulate pictures more intuitively. 
    We visualize the explainable and editable properties in HLS color space with control of the output with the simplest $OM_{add}$ operator mask. In Figure~\ref{fig:more_iharm}, we visualize the $OM_{add}$ in H, L, and S channels, and in Figure~\ref{fig:more_change}, we change the value of the $OM_{add}$ in each channel separately to see the editable attribute of the operator masks. We also binarize $OM_{add}$ with a threshold of 1e-4 and further calculate the mask IOU of $OM_{add}$ with ground truth masks (51.1\% for HCOCO, 85.3\% for HAdobe5k, 81.5\% for HFlickr, 77.9\% for Hday2night.)
    
\begin{table}[]
\centering

\caption{Quantitative results on four sub-datasets of iHarmony4. \textcolor{blue}{\textbf{Bold}} means the best results with our output format, \textbf{Blod} means the next best result. $^\ddag$ means we re-train the whole model.}
\resizebox{\columnwidth}{!}{%
\begin{tabular}{c|cc|cc|cc|cc|cc}
\hline
Sub-dataset & \multicolumn{2}{c|}{HCOCO} & \multicolumn{2}{c|}{HAdobe5k} & \multicolumn{2}{c|}{HFlickr} & \multicolumn{2}{c|}{Hday2night} & \multicolumn{2}{c}{All} \\ \hline
Evaluation Metric & MSE$\downarrow$ & PSNR$\uparrow$ & MSE$\downarrow$ & PSNR$\uparrow$ & MSE$\downarrow$ & PSNR$\uparrow$ & MSE$\downarrow$ & PSNR$\uparrow$ & MSE$\downarrow$ & PSNR$\uparrow$ \\ \hline
Composite & 69.37 & 33.94 & 345.54 & 28.16 & 264.35 & 28.32 & 109.65 & 34.01 & 172.47 & 31.63 \\
DoveNet(RGB)\cite{cong2020dovenet} & 36.72 & 35.93 & 52.32 & 34.34 & 133.14 & 30.21 & 54.05 & 35.18 & 52.36 & 34.75 \\
HT(RGB$^\ddag$)\cite{guo2021image} &  14.64&  39.04&  22.03&  38.85&  54.34&  34.12&  52.26&  36.75&  21.91 & 38.39 \\ \hline
HT(HLS, $OM_{add}$) & 21.60 & 37.45 & 29.48 & 36.89 & 78.29 & 32.59 & 63.81 & 36.36 & 31.00 & 36.73 \\
HT(HLS, $OM_{add}+OM_{mul}$) & 21.47 & 37.56 & 29.17 & 37.15 & 73.96 & 32.83 & 55.11 & 36.84 & 30.19 & 36.90 \\ 
HT(LAB, $OM_{add}+OM_{mul}$) & \textcolor{blue}{\textbf{13.95}} & \textcolor{blue}{\textbf{39.14}} & \textbf{22.04} & \textbf{38.64} & \textbf{59.03} & \textbf{34.00} & \textcolor{blue}{\textbf{43.57}} & \textcolor{blue}{\textbf{36.86}} & \textcolor{blue}{\textbf{21.89}} & \textbf{38.38} \\
\hline

\end{tabular}%
}

\label{tab:iharmony4}
\end{table}

\clearpage
%
%
\bibliographystyle{splncs04}
\bibliography{egbib}